\ifdefined\pdfminorversion
\fi
\documentclass[letterpaper, 10pt, conference]{ieeeconf}
\IEEEoverridecommandlockouts
\makeatletter
\let\NAT@parse\@undefined
\makeatother
\usepackage[numbers,sort&compress]{natbib} %
\usepackage{amsmath,amssymb,amsfonts}
\usepackage{algorithmic}
\usepackage{graphicx}
\usepackage{booktabs}
\usepackage{makecell}
\usepackage{tabularx}
\usepackage{cuted}
\makeatletter
\newcommand{\stripfigcaption}{\def\@captype{figure}\caption}
\newcommand{\striptabcaption}{\def\@captype{table}\caption}
\makeatother
\usepackage[section]{placeins}
\newcommand{\best}[1]{\underline{\textbf{#1}}}
\makeatletter
\ifdefined\@IEEEsectpunct \def\@IEEEsectpunct{.\ } \fi
\makeatother
\usepackage{textcomp}
\usepackage{xcolor}
\usepackage{url}
\usepackage[colorlinks=true,citecolor=blue,linkcolor=blue,urlcolor=blue]{hyperref}
\def\BibTeX{{\rm B\kern-.05em{\sc i\kern-.025em b}\kern-.08em
    T\kern-.1667em\lower.7ex\hbox{E}\kern-.125emX}}

\newcommand{\OURS}{RECAST}
\newcommand{\MAP}{AC map}

\begin{document}

\title{\vspace*{-20pt}\OURS{}: Recasting Vision-Language Semantics into \\ an Actionable Cost Map for Robot Navigation}

\author{Incheol Cho$^{1,3*}$, Jintae Park$^{2*}$, Jinkyu Kim$^{1,4}$,
Jungbeom Lee$^{1}$, Jaegul Choo$^{2\dagger}$, Seokha Moon$^{1\dagger}$\\[0.3em]
$^{1}$Korea University \quad $^{2}$KAIST AI \quad
$^{3}$Hanwha Aerospace \quad $^{4}$Kakao Mobility%
\thanks{$^{*}$Equal contribution.}%
\thanks{$^{\dagger}$Co-corresponding authors.}}

\maketitle

\begin{strip}
  \centering
  \includegraphics[width=.98\textwidth]{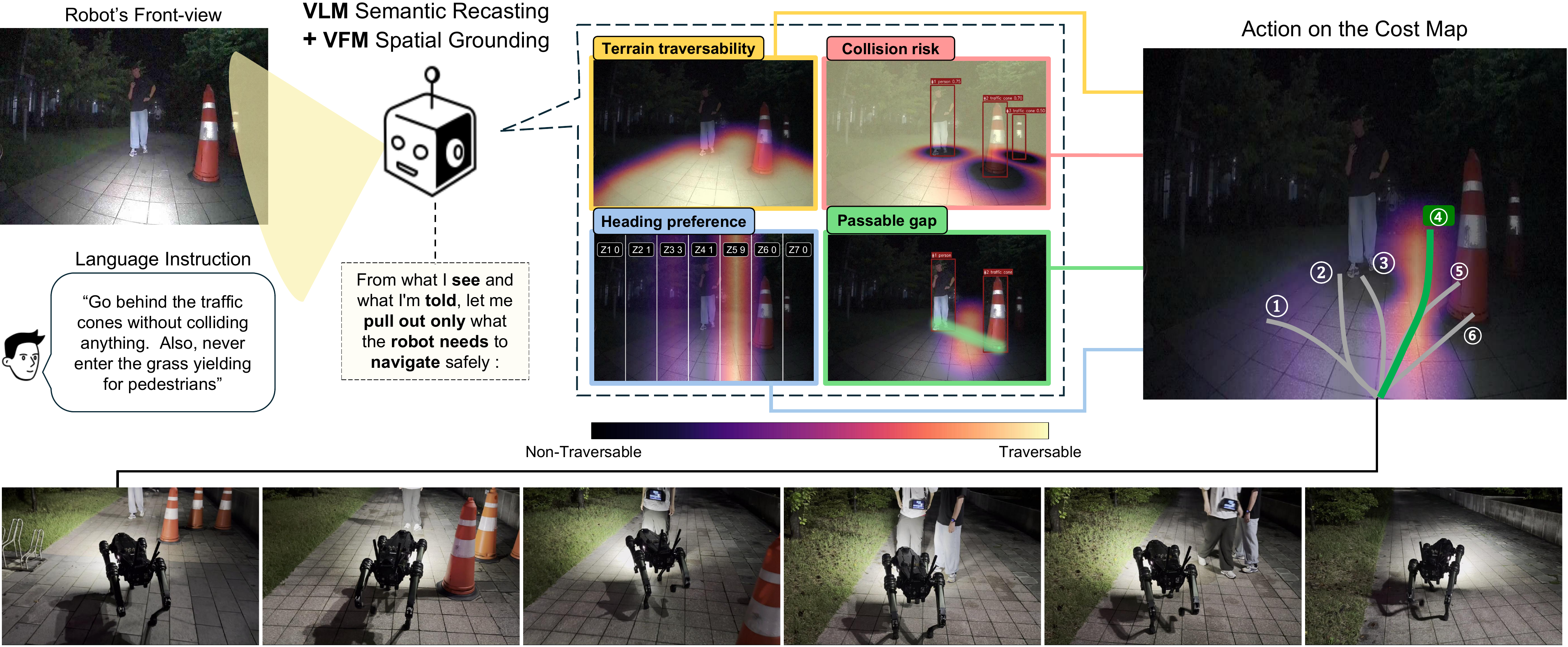}
  \vspace{-5pt}
  \stripfigcaption{Overview of \OURS{}. From the robot's front view and
      the user's instruction, a Vision-Language Model (VLM) judges terrain traversability, collision
      risk, heading preference, and passable gaps. Grounded in the
      image with Vision Foundation Models (VFMs), these judgments are recast
      into the Actionable Cost map. Candidate
      trajectories (gray) are scored on this map, and the best one (green)
      is executed. Below, the quadruped yields to the pedestrian, passes
      behind the traffic cones, and stays off the grass, as instructed.}  
  \label{fig:teaser}
\end{strip}

\begin{abstract}
Safe and robust robot navigation across diverse environments requires a high-level understanding of complex scenes and the ability to carry it into stable motion.
Recent works tackle this with learning-based models trained at scale and with approaches built on vision-language models (VLMs).
However, learning-based models break down outside their training distribution, while VLM-based approaches bring that understanding but rarely ground it in the scene or align the action with it.
To address these limitations, we present RECAST, a robot navigation framework that combines the reasoning of a VLM with the spatial grounding of vision foundation models to build an Actionable Cost map. 
Given the robot's front view and the user's instruction, we first decompose the scene with the VLM, judging which surfaces are traversable, which objects pose a risk, which heading to prefer, and which gaps are passable. 
Vision foundation models then ground these surfaces and objects in the image, and all four judgments are spatially recast into one compact cost map. 
This map both conditions the trajectory decoders and scores their proposals to select the one to execute. As the VLM's answers trail the live scene, both steps draw on cost maps from two points in time: the pivot frame the VLM judged, which carries all four judgments, and the current frame, whose terrain and collision costs are rebuilt from the current image.
RECAST improves success over the strongest prior method by 13.3 points in simulation and 31.4 points on a real quadruped, and reduces the collision rate relative to it by 9.6 and 14.3 points, reaching the lowest collision rate among all methods.
The project page is available at \url{https://recast-nav.github.io/}.
\end{abstract}

\section{Introduction}

Mobile robots are now part of everyday life, not only in indoor settings such as factories, retail stores, and hospitals, but also across diverse outdoor environments for last-mile delivery.
To move safely and robustly across such diverse
environments, a robot must build a
high-level understanding of a complex scene it has never seen and carry
that understanding through into stable motion. In Fig.~\ref{fig:teaser},
for example, the robot has to recognize a pedestrian to yield to, a gap
behind the traffic cones to thread, and pavement rather than grass to
stay on, and then move accordingly. And the scene keeps moving while the
robot deliberates, so that understanding must stay current as well.

These demands have driven a surge of research on robot navigation in two
directions. Learning-based navigation foundation
models~\cite{gnm2023,vint2023,nomad2024,exaug2023,crossemb2024,embagn2024,onering2024,navdp2025,se2025}
train on large-scale data to map observations directly to actions, and
run at high control rates across embodiments and environments. Approaches built on
vision-language models (VLMs) bring the high-level semantics those models lack, whether by
decoding actions from the reasoning and instruction following of a
language
model~\cite{navila2025,internvlan1,trackvla2025,navfom2025,ticvla2026,quarvla2024,quartonline2024,navid2024,uninavid2024,streamvln2025}
or by grounding the VLM's answers with vision foundation models into a
cost map or a trajectory choice~\cite{behav2025,convoi2024,vltgs2025}.

However, neither direction turns the high-level semantics of a scene
into accurate action. A policy trained on large-scale data often fails to cope with a scene
its data never covered, as benchmarks with diverse scenes
and moving pedestrians show~\cite{ticvla2026}. 
Where a VLM is added, its answer reaches action without being grounded in the scene,
inviting hallucination and reasoning-action
misalignment~\cite{dowhatyousay2025}. Even the approaches that ground
the instruction reduce the VLM's answer to one fixed cue, a rule
set~\cite{behav2025} or a trajectory pick~\cite{convoi2024,vltgs2025},
so the rest of what it reasons about the scene rarely shapes the action. What
is missing is a representation that recasts the high-level
semantics of a complex scene into a grounded, compact form that bears
directly on the action, and keeps that form aligned with the scene as it
moves.

To address these limitations, we introduce \textbf{\OURS{}}, a robot
navigation framework built around that missing representation
(Fig.~\ref{fig:teaser}). From the robot's egocentric view, the goal,
and the user's instruction, \OURS{} elicits from a VLM what in
the scene matters for moving through it, decomposed into four judgments:
(i) the traversability of terrain, (ii) the
collision risk each object poses, (iii) the heading to prefer, and (iv)
the passable gaps.
Vision foundation models then ground these judgments to their
corresponding regions, where they are laid out spatially and
composed into a single map that bears directly on the action, the
\textbf{Actionable Cost map} (\textbf{\MAP{}}). This map plays two
roles: it conditions the trajectory decoders, and it ranks their
trajectories to pick the one the robot follows, keeping the motion aligned with the reasoning. Since the scene
keeps changing while the VLM reasons, both roles use the map on two frames:
the pivot frame the VLM analyzed, holding all four judgments, and the
current frame, with terrain and collision costs refreshed per observation, so the map stays aligned
with the scene throughout the VLM's reasoning.

In summary, the contributions of our work are three-fold.
\begin{itemize}
\item We propose \OURS{}, a robot navigation framework that recasts the
high-level semantics of a VLM into the \MAP{}, which both conditions its
trajectory decoders and picks which of their trajectories to run.
\item We introduce a pivot--current design, which draws on the \MAP{}
of both the frame the VLM judged and the frame the robot currently sees, keeping
both conditioning and selection aligned with the live scene.
\item We achieve the highest success and the lowest collision rate
among all methods on a simulation benchmark
(DynaNav~\cite{ticvla2026}) and on a real quadruped robot.
\end{itemize}

\section{Related Works}

\subsection{Navigation Foundation Models}

Visual navigation has been learned end to end via imitation
\cite{codevilla2018,chaplot2018,diffdrive2025} and reinforcement learning \cite{ddpg2016,germ2024,diffdrivev2}, and extended to language instructions \cite{chaplot2018,wang2020} and open-vocabulary goals \cite{gervet2023,lelan2024}. Datasets collected across diverse embodiments and sensors \cite{scand2022,gnd2025,musohu2023,sacson2023,citywalker2025} have enabled navigation foundation models that generalize across embodiments and environments \cite{gnm2023,vint2023,nomad2024,exaug2023,crossemb2024,embagn2024,onering2024,navdp2025,se2025}. 
However, these policies carry only what their demonstrations
showed, so no new constraint can be added at deployment,
and a scene outside that data ends in collisions, as benchmarks with
moving pedestrians show~\cite{ticvla2026}.
In contrast, \OURS{} retains such learned decoders, but conditions them on the \MAP{} and additionally has the map evaluate their proposals, ensuring what the VLM knows reaches motion without retraining.

\subsection{VLM Reasoning for Navigation}

To handle out-of-distribution navigation scenes, many works leverage
VLM reasoning before predicting actions. Dual-system works pair a reasoning VLM with a low-level policy
\cite{navila2025,internvlan1,ticvla2026}, single-system vision-language-action models (VLAs) decode actions directly from video streams \cite{quarvla2024,quartonline2024,navid2024,uninavid2024,navida2026,streamvln2025,trackvla2025,navfom2025}, and others query a VLM at every step to select trajectories or motion
commands \cite{gptdriver2023,slowbrain2026,vlmsocialnav2025}.
Although remarkably powerful, these designs rarely ground the
reasoning to where it applies in the scene, so hallucinated or
misaligned reasoning~\cite{dowhatyousay2025} often leads to erroneous actions. 
BehAV~\cite{behav2025} does build a grounded cost map, but it
decomposes only the user's instruction into behavioral rules, so the
map covers the objects those rules name rather than what the scene
itself demands, and the goal enters its planner separately. Moreover, VLM reasoning takes seconds,
so even designs that pair it with a fast action decoder act on
answers that lag behind the dynamic scene. \OURS{} instead leverages the VLM to reason over
the complex scene itself and grounds each judgment, from terrain traversability to heading
preference and passable gaps,
into the \MAP{} before it reaches motion. Furthermore, its pivot--current design keeps this map aligned with the scene despite the slow reasoning.

\section{Methods}

\begin{figure*}[t]
  \centering
  \includegraphics[width=\textwidth, %
  ]{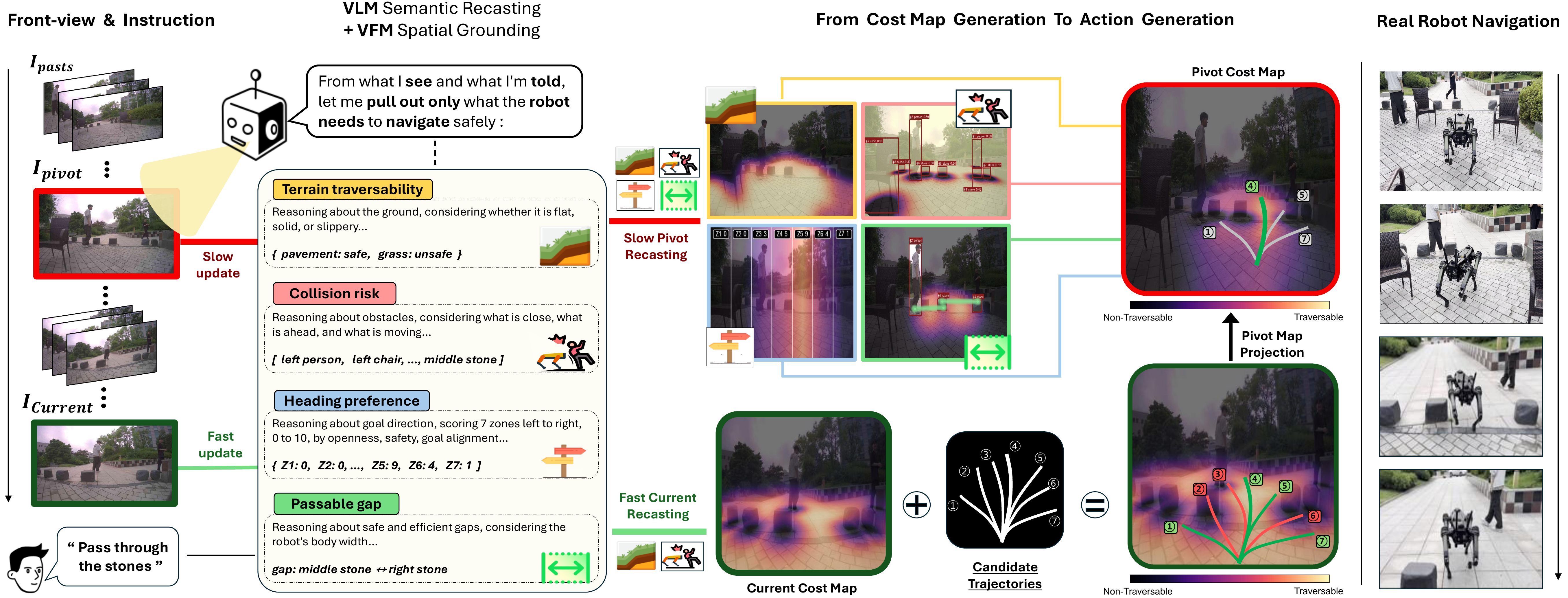}
  \vspace{-4pt}
  \caption{\OURS{} pipeline. Given the front view and the instruction, a
  VLM makes four judgments on the pivot frame, terrain traversability, collision
risk, heading preference, and passable gaps, which are
  grounded into four component cost maps and composed into the pivot
  \MAP{}. Each current frame rebuilds only the terrain traversability
  and collision risk components into the current \MAP{}, without a new
  VLM call.
  Conditioned on both maps, the decoder proposes candidate trajectories,
  each scored on the current map and, carried into the pivot frame, on
  the pivot map. The best one (green) is executed on the real robot
  (right), which passes through the gap between the stones as
  instructed.}
  \label{fig:arch}
  \vspace{-6pt}
\end{figure*}

\noindent\textbf{Preliminary.}
We consider instruction-guided point-goal navigation on a quadruped
robot. At control time $t$ the robot observes an egocentric RGB frame
$o_t$ together with a history $\mathcal{H}_t$ of past frames and the robot's past poses $\mathcal{P}_t$, both sampled at interval $\Delta$. It is further given a language instruction $\ell$ and a point goal. The policy then outputs $H$
future waypoints in the body frame, which a low-level controller
converts to velocity commands $(v,\omega)$.
Throughout, subscripts pv and cur tie a quantity to the \emph{pivot}
frame $o_{\mathrm{pv}}=o_{t-N\Delta}$, with an integer lag
$1\le N\le t/\Delta$, on which the latest VLM queries were issued,
or to the \emph{current} frame $o_{\mathrm{cur}}=o_t$, on which the
robot acts.

\subsection{The Actionable Cost Map (AC Map)}
\label{sec:vsmap}

What a VLM infers from a complex scene and the given instruction
steers the robot only once it is cast into a spatial form a policy
can act on.
The \MAP{} builds that form from VLM queries on a single frame
(Fig.~\ref{fig:arch}). The VLM answers them in labels, scores, and object
indices. Vision foundation models ground the surfaces and objects it names
to the pixels where they hold, while the direction and gaps follow
straight from its answers. The four stack as spatially
registered cost maps on the image plane, composed into one map.
Gaussian blurring keeps the map from splitting into discrete
regions, so cost changes gradually across the scene rather than
abruptly at object boundaries, and a path passing close to an obstacle
is penalized more than one passing far.

\smallskip\noindent\textbf{Terrain traversability cost map ($M^{T}$).}
The traversability query carries a prompt-side guide of the objects and
terrain to watch for, together with the constraints the user states.
The prompted VLM reasons over the whole scene, naming the objects
and surfaces in view and judging how traversable each is, then
hands the scored labels to a text-conditioned segmentation
model~\cite{clipseg}, whose per-label probability maps label each
pixel, and a Gaussian kernel smooths the cost map,
\begin{equation}
  M^{T} \;=\; G_{\sigma} * s_{c^{\star}}, \qquad
  c^{\star}=\arg\max_{c} P_c ,
\label{eq:layerA}
\end{equation}
where $P_c$ is the segmentation probability of label $c$ and
$s_c\in[0,1]$ its traversability. Higher values mean preferred ground
(e.g.\ sidewalk over gravel), lower values obstacles and hazards, and
a constraint stated in language changes $s_c$ directly.

\smallskip\noindent\textbf{Collision risk cost map ($M^{C}$).}
Every object the traversability query flags as an obstacle is located by an
open-vocabulary detector~\cite{yoloworld} and punches a graded
Gaussian hole in the map at the ground point of its box, with
metric radius $R$ and depth one minus its traversability $s_i$,
and overlapping holes keep the deepest value. An obstacle the
instruction $\ell$ drove to zero is thus cut out entirely, which
turns a constraint stated in language into geometry, and $R$ can be
enlarged at inference without retraining.

\smallskip\noindent\textbf{Heading preference cost map ($M^{H}$).}
The heading query partitions the frame into seven vertical zones and
scores each one given the instruction $\ell$ and the goal bearing, so a
zone scores high only when it is both oriented toward the goal and
traversable. The direction is thus the VLM's own spatial judgment. The
scores, given on a 0--10 scale and normalized to $[0,1]$, are
interpolated across image columns into a horizontal preference field, and this is how the intent to reach the goal enters
the map.

\smallskip\noindent\textbf{Passable gap cost map ($M^{P}$).}
The passable gap query names object \emph{pairs} whose gap is passable. Each
passable pair opens a Gaussian corridor between the two footprints, a
column centered on the gap whose width follows the gap. $M^{P}$ equals
one everywhere else and rises toward $\gamma$ inside the corridor, so
whereas the other three only attenuate, $M^{P}$ alone boosts, restoring narrow passages that
$M^{C}$ would otherwise close.

\smallskip\noindent\textbf{Composition.}
The four components are composed into a single compact image-space map,
\begin{equation}
  M_{\mathrm{in}} = \big[\,
     \gamma^{-1} M^{T}\!\odot\!(1{-}M^{C})\!\odot\!
     M^{H}\!\odot\!M^{P}\,\big]_{0}^{1},
\label{eq:min}
\end{equation}
where $M^{T}$, $M^{C}$, and $M^{H}$ lie in $[0,1]$, $M^{P}$ in
$[1,\gamma]$, and $[\cdot]_{0}^{1}$ clamps the composite to $[0,1]$
after the gap boost. The policy consumes $M_{\mathrm{in}}$ as a dense
conditioning input. The composite is built on two frames (Fig.~\ref{fig:arch}).
All four components are
baked on the pivot frame, the one where the VLM answers are
available, giving $M_{\mathrm{in},\mathrm{pv}}$. On every current frame it is rebuilt from $M^{T}$ and $M^{C}$ alone,
reusing the labels and scores the VLM returned on the pivot frame and
rerunning only the segmenter and the detector on the current frame,
which gives $M_{\mathrm{in},\mathrm{cur}}$ without a new VLM call. What the current
frame refreshes is therefore where things are, not what they mean.

\subsection{AC Map-Conditioned Trajectory Decoders}
\label{sec:decoders}

The \MAP{} is decoder-agnostic. We instantiate two trained decoders
that consume it through the pivot--current conditioning described
below.
Following anchor-based multi-modal planners~\cite{diffdrive2025}, both
cluster the training trajectories with $K$-means into
$K$ anchors $\{a^{(k)}\}_{k=1}^{K}$ and predict one trajectory per
anchor, so the multi-modality of feasible paths is preserved by
construction.

\smallskip\noindent\textbf{Pivot--current conditioning.}
\label{sec:pivot}
VLM judgment and generated action sit on different frames. The decoder therefore
takes both instances of the map, $M_{\mathrm{in},\mathrm{pv}}$ and
$M_{\mathrm{in},\mathrm{cur}}$, as separate streams, and is told the
lag $N$ between them.
Each anchor owns one trajectory query, and the $K$ queries $Q$, built
from the past poses
$\mathcal{P}_t$, pass through self-attention among themselves and
cross-attention over four token streams, one from the past frames,
two from the pivot frame, and one from the current frame. The
\emph{history} stream holds one token per frame of $\mathcal{H}_t$.
The \emph{pivot object} stream $O_{\mathrm{pv}}$ holds one token per
object, and per passable pair, that the VLM returned on the pivot
frame. The
\emph{pivot semantic} stream $S_{\mathrm{pv}}$ fuses
$o_{\mathrm{pv}}$ with $M_{\mathrm{in},\mathrm{pv}}$, and the
\emph{current semantic} stream $S_{\mathrm{cur}}$ fuses
$o_{\mathrm{cur}}$ with $M_{\mathrm{in},\mathrm{cur}}$, each by
concatenating, cell by cell, the features of a frozen image encoder
$\phi$~\cite{clip} and of a small map encoder $\psi$, so every token
carries both appearance and the map. The decoder maps the queries and the
four streams to $K$ trajectories and their scores,
\begin{equation}
  \{\hat\tau^{(k)},p^{(k)}\}_{k=1}^{K} =
    \pi_{\theta}\big(Q;\;\mathcal{H}_t,\,O_{\mathrm{pv}},\,
    S_{\mathrm{pv}},\,S_{\mathrm{cur}},\,e_N\big),
\label{eq:policy}
\end{equation}
where $\hat\tau^{(k)}$ is the trajectory predicted for mode $k$ and
$p^{(k)}$ the confidence a classification head assigns to it. The lag
$N$ enters twice, as a one-hot token $e_N$ on the pivot semantic
stream and as a marker on the pivot's slot in the history stream, so
the history slots between the pivot and the present are exactly what
changed between reasoning and control. Sampling $N$ uniformly during
training lets one model span every delay. The decoders never receive
the goal directly. It reaches them only through the \MAP{}, where
$M^{H}$ encodes it.

\smallskip\noindent\textbf{Training objective.}
Each sample supervises only the mode whose anchor is nearest to its
ground-truth trajectory,
$k^{\mathrm{gt}}=\arg\min_{k}\|a^{(k)}-\tau^{\mathrm{gt}}\|_{2}$. Let
$\hat W$ be the cumulative waypoints of the predicted trajectory
$\hat\tau^{(k^{\mathrm{gt}})}$ of that mode, $W^{\mathrm{gt}}$ those of
the ground truth, and $J(\cdot)$ the jerk of a waypoint sequence.
The loss sums an $L_1$ trajectory term, an $L_1$ jerk term, and a
cross-entropy term,
\begin{equation}
\begin{aligned}
  \mathcal{L} =\;&\big\|\hat W-W^{\mathrm{gt}}\big\|_{1}
     +\lambda\big\|J(\hat W)-J(W^{\mathrm{gt}})\big\|_{1}\\
    &+\mathrm{CE}\big(p,k^{\mathrm{gt}}\big),
\end{aligned}
\label{eq:loss}
\end{equation}
with $\lambda$ weighting the jerk term and $p$ the score of the
classification head. We add the jerk term because the trajectory term alone does not
penalize a path that zig-zags between waypoints.

\smallskip\noindent\textbf{Decoder variants.}
The two trained decoders, each trained as its own model, differ in
how $Q$ is formed.
The regression decoder projects the embedding of
$\mathcal{P}_t$ into $K$ queries, evaluates $\pi_\theta$ once, and
predicts a per-anchor residual, $\hat\tau^{(k)}=a^{(k)}+\delta^{(k)}$.
The diffusion decoder uses truncated diffusion~\cite{diffdrive2025},
forming each query from an anchor noised up to a truncation step plus
the same embedding, and evaluating $\pi_\theta$ once per DDIM step to
denoise it back, so
$\hat\tau^{(k)}=\hat x_0^{(k)}$.

\subsection{AC Map-Scored Mode Selection}
\label{sec:select}

The trained decoders emit $K$ candidates and leave the choice to their
classification head, which picks the mode with the highest confidence
$p^{(k)}$. But that head only imitates which anchor the
demonstration took. It never scores a path for progress toward a
goal, for safety, or for whether the ground it crosses was judged
traversable. No goal appears in its training data or in any of its
inputs. And being trained, its judgment does not carry beyond that
distribution. The \MAP{} is exactly those judgments
in spatial form, so \OURS{} hands the choice to the map, projecting each
candidate onto it and scoring the ground the trajectory crosses.

\smallskip\noindent\textbf{Scoring map.}
The components are rasterized onto a metric bird's-eye-view grid by
ground-plane projection,
\begin{equation}
  M_{\mathrm{score}} = \operatorname{BEV}\!\big(M^{T}\!\odot\!M^{H}\!\odot\!M^{P}\big)
     \!\odot\!(1{-}\operatorname{BEV}(M^{C})),
\label{eq:mscore}
\end{equation}
with $M^{C}$ applied after rasterization so its holes stay
metric. Like $M_{\mathrm{in}}$, it exists on both frames,
$M_{\mathrm{score},\mathrm{pv}}$ from all four components and
$M_{\mathrm{score},\mathrm{cur}}$ from $M^{T}$ and $M^{C}$.

\smallskip\noindent\textbf{Mode selection and safety veto.}
The mode that is executed is the one with the highest score,
\begin{equation}
  k^{\star}=\arg\max_{k}\; s^{(k)},
\label{eq:sel}
\end{equation}
and its trajectory $\hat\tau^{(k^{\star})}$ is what the low-level
controller tracks. The score carries the safety veto, which zeroes any
candidate whose current-frame term falls below $\tau$ and stops the
robot if no mode clears it,
\begin{equation}
  s^{(k)} = \begin{cases}
      0 & (r_{\mathrm{cur}} < \tau), \\
      r_{\mathrm{cur}}\, r_{\mathrm{pv}}\, r_{\mathrm{align}}
        & (r_{\mathrm{cur}} \ge \tau),
    \end{cases}
\label{eq:gate}
\end{equation}
where every factor is evaluated on candidate $k$.

\smallskip\noindent\textbf{Score terms.}
Let $x_{h}$ be waypoint $h$ of candidate $k$ in the current body frame
and $\tilde{x}_{h}$ the same
waypoint carried into the pivot frame by the accumulated SE(2) odometry.
A waypoint is \emph{trusted} in a map when its ground cell lies inside
that camera's ground coverage. A candidate with no trusted waypoint
scores zero, so unseen ground counts as unsafe. Writing
$\mathcal{T}_{\mathrm{cur}}$ for the waypoints trusted by
$M_{\mathrm{score},\mathrm{cur}}$ inside a forward window and
$\mathcal{T}_{\mathrm{pv}}$ for those trusted by
$M_{\mathrm{score},\mathrm{pv}}$, the three factors are
\begin{equation}
\begin{aligned}
  r_{\mathrm{cur}} &= P_{q}\big\{\,M_{\mathrm{score},\mathrm{cur}}(x_{h})\;:\;
      h\in\mathcal{T}_{\mathrm{cur}}\big\}, \\
  r_{\mathrm{pv}} &= \textstyle\sum_{h\in\mathcal{T}_{\mathrm{pv}}} h\,
      M_{\mathrm{score},\mathrm{pv}}(\tilde{x}_{h})
      \,\big/\, \textstyle\sum_{h\in\mathcal{T}_{\mathrm{pv}}} h, \\
  r_{\mathrm{align}} &= \exp\!\big(-w\,|\Delta\theta|/\sigma_{\theta}\big),
\end{aligned}
\label{eq:terms}
\end{equation}
where $P_{q}$ is the $q$-th percentile and $\Delta\theta$ is the
angle between the goal bearing and the heading of the candidate.
$r_{\mathrm{cur}}$ pools by a low percentile, so the veto rejects
any candidate whose riskiest trusted samples are unsafe. $r_{\mathrm{pv}}$ averages the pivot map along the
candidate with weights that grow linearly toward the last waypoint,
so it cares most about where the trajectory ends up. All three
factors rank the candidates that pass, but only
$M_{\mathrm{score},\mathrm{cur}}$ can reject one, so a stale pivot
never admits an unsafe path.

The \MAP{} thus governs both what the decoders \emph{propose}
(Eq.~\ref{eq:min}) and what the robot finally \emph{executes}
(Eq.~\ref{eq:mscore} and~\ref{eq:sel}).
It is also actionable with no learned decoder at all. A metric-depth
network~\cite{dav2} lifts the pivot and current maps into a world BEV
cost grid, and a sampling-based MPC optimizes unicycle rollouts on it
and returns one trajectory, with no trained component in the loop.

\section{Experiments}
\label{sec:exp}

\subsection{Experimental Setup}
\label{sec:impl}

\smallskip\noindent\textbf{Datasets.}
Both decoders are trained on a mixture of
SCAND~\cite{scand2022}, GND~\cite{gnd2025}, and our teleoperated
Vision60 dataset. Evaluation spans simulation and the real world. In
simulation we follow the official 85-episode
DynaNav protocol~\cite{ticvla2026}, 25 episodes each in a hospital,
an office, and a warehouse, and 10 outdoors with moving pedestrians,
replacing only its wheeled robot with a Spot quadruped for every
method. In the real world we deploy on a Vision60 quadruped over
seven campus scenes, three indoor and four outdoor, each run five
times with goals 9 to 16\,m away.

\smallskip\noindent\textbf{Implementation details.}
The VLM is Gemini~3.5 Flash-Lite, and its answers are grounded by
CLIPSeg~\cite{clipseg} for surfaces and YOLO-World~\cite{yoloworld}
for objects. Each \MAP{} component is rendered at $224{\times}224$
with $\gamma{=}1.3$ and $R{=}0.25$\,m ($0.4$\,m at deployment), and
the BEV grid covers $12{\times}10$\,m at 0.05\,m per cell. Both
decoders use $K{=}10$ anchors, $H{=}10$ waypoints at
$\Delta{=}0.5$\,s, and a frozen CLIP ViT-B/16~\cite{clip} as $\phi$,
and the diffusion decoder denoises in two DDIM steps from
$T_{\mathrm{trunc}}{=}200$. They are trained for 20 epochs on one
H200 GPU with AdamW (batch 1024, learning rate $2{\times}10^{-4}$,
$\lambda{=}0.5$), with $N$ drawn from $\{2,\dots,10\}$ (1--5\,s) and
clamped to $\{2,\dots,6\}$ (1--3\,s) at deployment. Mode selection
uses $\tau{=}0.35$ and a 1--4\,m window for $r_{\mathrm{cur}}$ in
simulation ($0.15$ and a 3\,m window on the real robot), $q{=}10$,
and $w{=}1$ and $\sigma_{\theta}{=}45^{\circ}$ for
$r_{\mathrm{align}}$. Simulation runs on an RTX~3090 and the real
robot on a laptop RTX~5080 over ROS2.

\smallskip\noindent\textbf{Evaluation metrics.}
We report the benchmark's official metrics~\cite{ticvla2026}, success
rate (SR) within 1.5\,m of the goal before timeout, navigation error
(NE) in meters, success weighted by path length (SPL), and collision
rate (CR), counting a sustained contact or a pass within 0.2\,m of a
person.
In the real world the same success criterion is checked against UTM
tracks outdoors and recorded video indoors.

\subsection{Comparison with Prior Methods}

\begin{table}[t]
\centering
\caption{Results on the DynaNav benchmark~\cite{ticvla2026}. NavDP$^{\dagger}$ receives the point goal, NavDP does not. Bold with underline marks the best in each column.}
\label{tab:sim}
\vspace{-4pt}
\footnotesize
\setlength{\tabcolsep}{2pt}
\renewcommand{\arraystretch}{1.15}
\begin{tabular*}{\columnwidth}{@{\extracolsep{\fill}}l cccc@{}}
\toprule
\textbf{Method} & \textbf{SR(\%)}\,$\uparrow$ & \textbf{NE(m)}\,$\downarrow$ & \textbf{SPL(\%)}\,$\uparrow$ & \textbf{CR(\%)}\,$\downarrow$ \\
\midrule
ViNT \cite{vint2023}                & 10.6 & 21.2 & \phantom{0}9.7 & 64.1 \\
NoMaD \cite{nomad2024}              & 30.0 & 15.2 & 24.9 & 35.3 \\
NaVIDA \cite{navida2026}            & 22.4 & 18.3 & 19.1 & 30.0 \\
TIC-VLA \cite{ticvla2026}           & 35.9 & 12.6 & 32.6 & 31.2 \\
NavDP \cite{navdp2025}              & 21.8 & 21.3 & 19.9 & 24.7 \\
NavDP$^{\dagger}$ \cite{navdp2025}  & 66.7 & \phantom{0}5.80 & 64.0 & 32.0 \\
\midrule
\OURS{} (MPC)         & 71.8 & \phantom{0}5.84 & 66.8 & 30.6 \\
\textbf{\OURS{} (regression)}               & 77.6 & \phantom{0}\best{4.07} & 68.6 & \best{22.4} \\
\textbf{\OURS{} (diffusion)}       & \best{80.0} & \phantom{0}4.73 & \best{70.0} & 24.7 \\
\bottomrule
\end{tabular*}
\vspace{-4pt}
\end{table}

\begin{table}[t]
\centering
\caption{Real-world results on the Vision60 quadruped, seven campus scenes with five runs each. Bold with underline marks the best in each column.}
\label{tab:real}
\vspace{-4pt}
\footnotesize
\setlength{\tabcolsep}{2pt}
\renewcommand{\arraystretch}{1.15}
\begin{tabular*}{\columnwidth}{@{\extracolsep{\fill}}l cccc@{}}
\toprule
\textbf{Method} & \textbf{SR(\%)}\,$\uparrow$ & \textbf{NE(m)}\,$\downarrow$ & \textbf{SPL(\%)}\,$\uparrow$ & \textbf{CR(\%)}\,$\downarrow$ \\
\midrule
ViNT \cite{vint2023}      & 22.9 & \phantom{0}8.07 & 19.3 & 54.3 \\
NoMaD \cite{nomad2024}    & 45.7 & \phantom{0}3.90 & 33.6 & 42.9 \\
NavDP \cite{navdp2025}    & 31.4 & \phantom{0}7.45 & 22.8 & 54.3 \\
NaVIDA \cite{navida2026}  & \phantom{0}5.7 & 11.12 & \phantom{0}4.6 & 37.1 \\
TIC-VLA \cite{ticvla2026} & 22.9 & \phantom{0}8.29 & 18.9 & 45.7 \\
\midrule
\textbf{\OURS{}} & \best{77.1} & \phantom{0}\best{1.94} & \best{60.2} & \best{28.6} \\
\bottomrule
\end{tabular*}
\vspace{-4pt}
\end{table}

\begin{table}[t]
\centering
\newcommand{\cmk}{$\checkmark$}
\newcommand{\gr}[1]{\textcolor{black!40}{#1}}
\caption{Ablation of the \MAP{} components and the pivot--current design with the regression decoder. pv: pivot cost map, cur: current cost map.}
\label{tab:ablation}
\vspace{-4pt}
{\footnotesize\setlength{\tabcolsep}{1pt}\renewcommand{\arraystretch}{1.15}%
\begin{tabular*}{\columnwidth}{@{\extracolsep{\fill}}l cccc cc cccc@{}}
\toprule
 & \multicolumn{4}{c}{\textbf{Map components}} & \multicolumn{2}{c}{\textbf{Frames}} & \multicolumn{4}{c}{\textbf{DynaNav}} \\
\cmidrule(lr){2-5}\cmidrule(lr){6-7}\cmidrule(lr){8-11}
\textbf{Variant} & $M^{T}$ & $M^{C}$ & $M^{H}$ & $M^{P}$ & pv & cur & \textbf{SR}\,$\uparrow$ & \textbf{NE}\,$\downarrow$ & \textbf{SPL}\,$\uparrow$ & \textbf{CR}\,$\downarrow$ \\
\midrule
\textbf{\OURS{} (ours)} & \cmk & \cmk & \cmk & \cmk & \cmk & \cmk & \textbf{77.6} & \textbf{4.07} & \textbf{68.6} & \textbf{22.4} \\
\midrule
Model A & \cmk &  &  &  & \gr{\cmk} & \gr{\cmk} & 58.8 & 7.59 & 53.8 & 35.3 \\
Model B & \cmk & \cmk &  &  & \gr{\cmk} & \gr{\cmk} & 69.4 & 5.70 & 64.0 & 34.1 \\
Model C & \cmk & \cmk & \cmk &  & \gr{\cmk} & \gr{\cmk} & 74.1 & 4.97 & 68.0 & \textbf{22.4} \\
\midrule
Model D & \gr{\cmk} & \gr{\cmk} & \gr{\cmk} & \gr{\cmk} & \cmk &  & 65.9 & 5.91 & 59.5 & 32.9 \\
\bottomrule
\end{tabular*}}
\medskip
\caption{Ablation of the roles the \MAP{} serves with the regression decoder.}
\label{tab:roles}
\vspace{-4pt}
{\footnotesize\setlength{\tabcolsep}{1pt}\renewcommand{\arraystretch}{1.15}%
\begin{tabular*}{\columnwidth}{@{\extracolsep{\fill}}l cc c cccc@{}}
\toprule
 & \multicolumn{2}{c}{\textbf{Map role}} & \textbf{Final choice} & \multicolumn{4}{c}{\textbf{DynaNav}} \\
\cmidrule(lr){2-3}\cmidrule(lr){4-4}\cmidrule(lr){5-8}
\textbf{Variant} & input & scoring & selector & \textbf{SR}\,$\uparrow$ & \textbf{NE}\,$\downarrow$ & \textbf{SPL}\,$\uparrow$ & \textbf{CR}\,$\downarrow$ \\
\midrule
\textbf{\OURS{} (ours)} & full & full & map & \textbf{77.6} & \textbf{4.07} & \textbf{68.6} & \textbf{22.4} \\
\midrule
Model A & $M^{T}$ & \gr{full} & \gr{map} & 64.7 & 6.34 & 59.6 & 31.8 \\
Model B & \gr{full} & $M^{T}$ & \gr{map} & 64.7 & 6.49 & 59.0 & 28.2 \\
\midrule
Model C & \gr{full} & \gr{full} & head & 14.1 & 18.25 & 12.7 & 29.4 \\
\bottomrule
\end{tabular*}}
\vspace{-4pt}
\end{table}

\begin{figure*}[t]
  \centering
  \includegraphics[width=\textwidth]{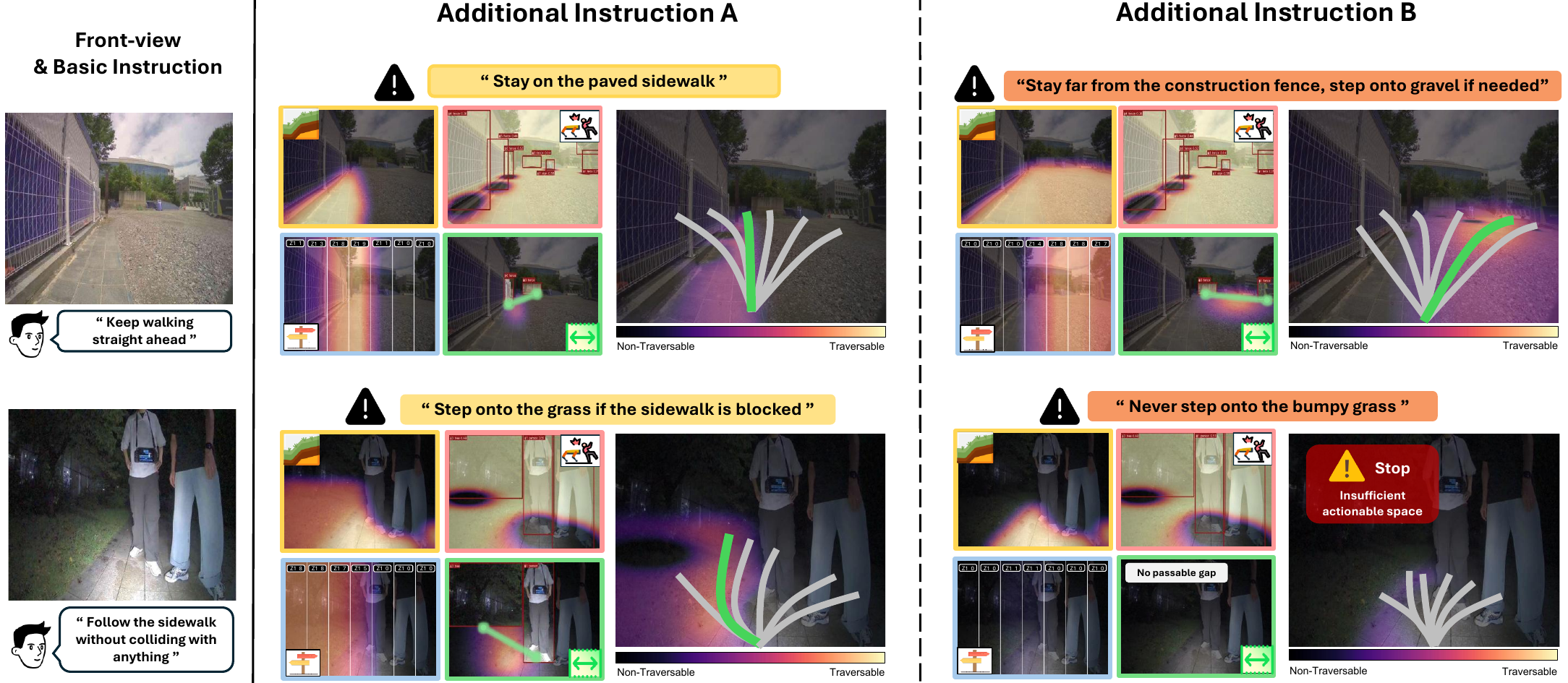}
  \vspace{-4pt}
  \caption{The \MAP{} and how an added instruction reshapes it. Two real
  scenes, each under a basic instruction (left) and two additional
  instructions, A and B. Each panel shows the four components as insets
  and the composed \MAP{} overlaid on the frame, with the $K$
  candidate trajectories in gray and the selected one in green. A
  panel marked Stop is a safety veto, with every candidate rejected.
  All runs use the regression decoder.}
  \label{fig:castmap}
  \vspace{-6pt}
\end{figure*}

\smallskip\noindent\textbf{Simulation.}
The three \OURS{} variants in Table~\ref{tab:sim} share the same
\MAP{} and differ only in what acts on it, a training-free MPC, a
regression decoder, or a diffusion decoder, and all three achieve a
higher SR than every prior method. With the MPC alone, \OURS{} reaches
71.8\% SR, surpassing the strongest prior method, NavDP$^{\dagger}$
(66.7\%), which receives the same point goal, so the map is enough
to guide the robot even without a learned decoder. Plugging in the
trained decoders raises SR further, to 77.6\% with regression and
80.0\% with diffusion, and each favors different metrics. The
regression decoder achieves the lowest NE (4.07\,m) and CR (22.4\%)
of all methods, while the diffusion decoder attains the highest SPL
(70.0\%) along with its top SR. The decoder can thus be swapped to
favor safety or success while the map stays unchanged.

\smallskip\noindent\textbf{Real world.}
Table~\ref{tab:real} compares all methods on the real Vision60
quadruped. Running its regression decoder, \OURS{} leads on every
metric, with 77.1\% SR, 1.94\,m NE, 60.2\% SPL, and the lowest CR of
28.6\%. The strongest prior method, NoMaD, reaches 45.7\% SR but collides
in 42.9\% of runs, as a policy learned from demonstrations alone does
not reflect what the scene or the instruction demands. The VLM-based
prior methods, TIC-VLA and NaVIDA, take the instruction yet reach only
22.9\% and 5.7\% SR, which we attribute to reasoning that is not
grounded in the scene. \OURS{} grounds each judgment in the \MAP{}
before it reaches motion, and it ends closer to the goal on average,
with an NE less than half that of NoMaD.
\subsection{Ablations}

\smallskip\noindent\textbf{Effect of the \MAP{} components.}
Table~\ref{tab:ablation} stacks the components of the \MAP{} one by
one. Model A keeps only the terrain traversability cost map $M^{T}$, which is what
scene-decomposition approaches such as BehAV~\cite{behav2025} act
on, one instruction-conditioned cost map, and it reaches 58.8\% SR
with 35.3\% CR.
BehAV itself plans on 3D LiDAR, so Model A stands in as its
camera-only counterpart. Decomposing the vision-language semantics
further pays at every step. The collision risk map $M^{C}$ (Model B) adds 10.6 points
of success, the heading preference $M^{H}$ (Model C) adds another 4.7 while
cutting collisions from 34.1\% to 22.4\%, and the passable gap
map $M^{P}$ adds a final 3.5 by reopening the narrow
passages $M^{C}$ closes. The full map improves on Model A by
18.8 points of success and 12.9 points of collision rate, every
metric moving in the same direction. Each component carries a judgment
the previous ones cannot express, so the gain comes from decomposing
further, not from sharpening one map.

\smallskip\noindent\textbf{Pivot--current design.}
Model D in Table~\ref{tab:ablation} is a single-frame variant, retrained to assume pivot equals current,
removing the current semantic stream and the lag token from the
decoder and moving the veto onto the pivot map. Every metric
degrades, success falls from 77.6\% to 65.9\%, NE grows from 4.07\,m to 5.91\,m, SPL drops from 68.6\% to 59.5\%, and CR rises from 22.4\% to 32.9\%. The variant still navigates, so
the delayed judgments alone keep the system viable, but planning and
vetoing on a scene that is seconds old costs accuracy, path quality,
and safety at once. The current frame's refresh of where things are
is what buys them back.

\smallskip\noindent\textbf{\MAP{}-scored mode selection.}
The map serves two roles, the decoders' input and the scoring
space (Table~\ref{tab:roles}). Models A and B replace the full map with $M^{T}$ in the input or the scoring alone, which drops
success from 77.6\% to 64.7\% and raises collisions from 22.4\% to 31.8\% and 28.2\%, so each role needs the full map and neither compensates
for the other. Model C hands the final choice to the classification head instead,
with the same candidates and the same veto, which collapses success to 14.1\%, since the confidence $p$ only imitates the demonstrations and
carries no notion of goal, safety, or traversability
(Sec.~\ref{sec:select}). Conditioning the decoder on the map is thus not enough. The map must also make the final choice.

\subsection{Qualitative Analysis}

\begin{figure}[t]
  \centering
  \includegraphics[width=\columnwidth]{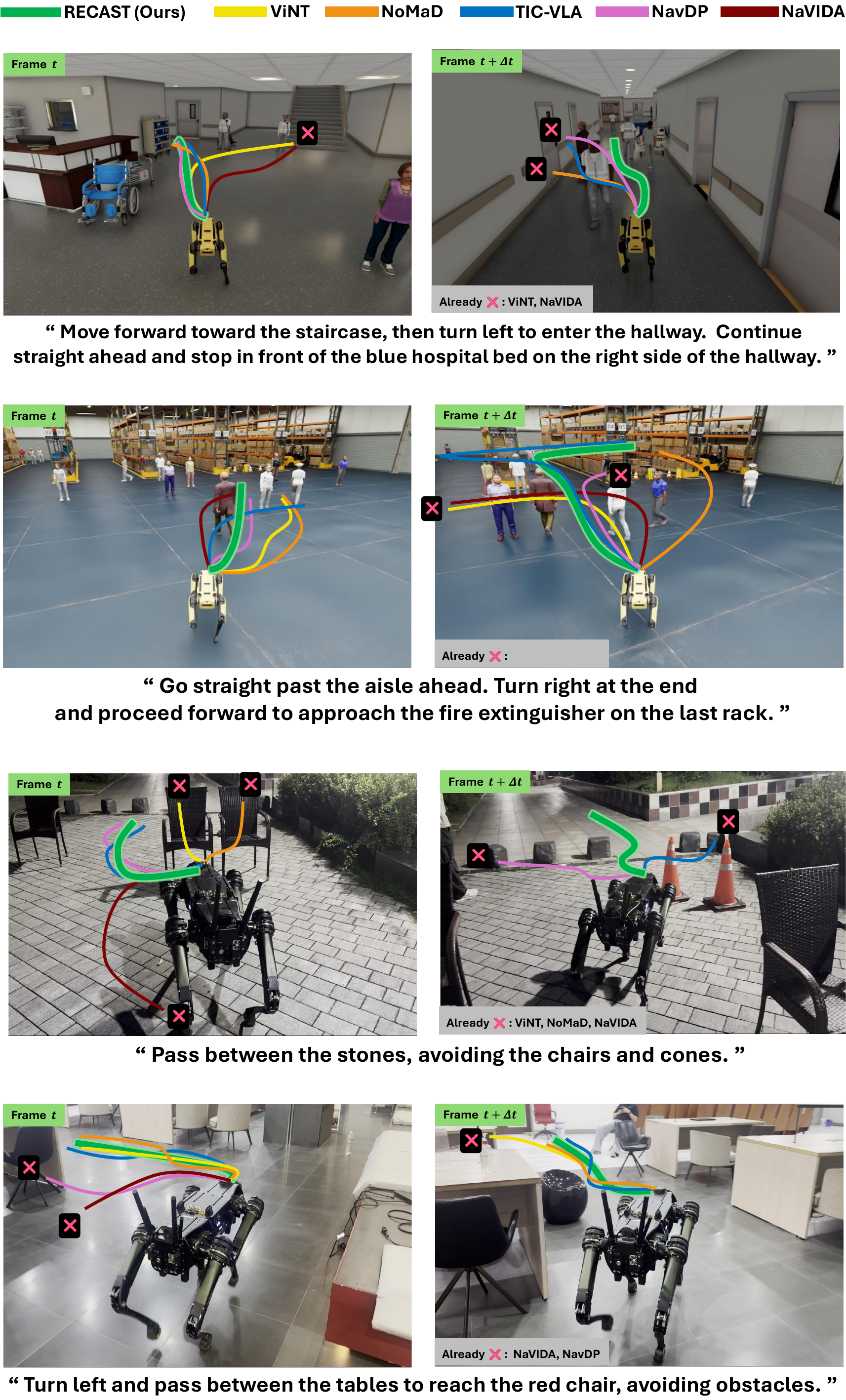}
  \vspace{-15pt}
  \caption{Qualitative results. Two DynaNav episodes (top), a
  hospital hallway and a warehouse aisle, and two real-world runs on
  the Vision60 (bottom), passing between curb stones among chairs and cones, and between tables toward a named chair. Each is
  shown at a frame $t$ and a later frame $t{+}\Delta t$ with the
  trajectories of all six methods overlaid on the robot's view and
  the instruction below. Methods that have already failed by
  $t{+}\Delta t$ are listed in the panel. \OURS{} runs its regression
  decoder.}
  \label{fig:quali}
  \vspace{-6pt}
\end{figure}

\smallskip\noindent\textbf{Effect of the instruction on the \MAP{}.}
Fig.~\ref{fig:castmap} shows how one added sentence changes the
map. Two real scenes are each run under a basic instruction and two
additional ones. At the construction fence, ``stay on the paved
sidewalk'' keeps the sidewalk bright and the straight candidate
wins, whereas ``stay far from the construction fence, step onto
gravel if needed'' widens the hole around the fence and the turn onto
the gravel wins. On the night sidewalk, ``step onto the grass if the
sidewalk is blocked'' lights the grass and a left turn across it is
selected, whereas ``never step onto the bumpy grass'' darkens the
grass, so with the pavement blocked by the person every candidate
crosses the grass or the hole around the person, the safety veto
rejects them all, and the robot stops. The constraint changed only the map, and
the map changed the path, or stopped the robot.

\smallskip\noindent\textbf{Simulation.}
The top two rows of Fig.~\ref{fig:quali} follow all six methods
through a hospital and a warehouse episode. In the hospital,
pedestrians walk toward the robot once it turns into the hallway, and
the current-frame veto drops every mode that would meet them.
\OURS{} bends between them and comes out clear on the other side,
while ViNT and NaVIDA have already failed by $t{+}\Delta t$. In the
warehouse, workers crowd the end of an aisle lined with stacked
racks, which a quadruped cannot bump without risking a fall. \OURS{} threads the gap between the workers
into the right turn without touching a rack, whereas the prior methods cut
through the crowd or swing wide.

\smallskip\noindent\textbf{Real world.}
The bottom two rows of Fig.~\ref{fig:quali} show two campus runs.
Outdoors at night, at the curb stones, \OURS{} steers clear of the
chairs and cones and passes between the stones as instructed. Indoors,
it turns left, skirts a dark, oddly shaped chair in its path, passes
between the tables, and stops at the red chair. By $t{+}\Delta t$,
three prior methods have already failed in the outdoor run and two in the
indoor run, and the rest, which either cannot
take the instruction or do not follow it, wander off across the open
floor.

\makeatletter\def\@fb@secFB{}\makeatother
\section{Conclusion}

We presented \OURS{}, a robot navigation framework that recasts the
scene judgments of a VLM, grounded by vision foundation models, into
the \MAP{}. The map both conditions the trajectory decoders and selects
which of their proposals to execute, and, built on both the pivot
and current frames, stays aligned with the live scene while the
VLM reasons. It achieves the highest success and lowest
collision rate in simulation and the real world, and a new
instruction reshapes the path through the map alone.
Its main limitation lies in the grounding models, since whatever they
miss never enters the map. Future work will predict obstacle motion to
bridge the gap between the pivot and current frames.


\begin{thebibliography}{00}

\bibitem{gnm2023} D. Shah, A. Sridhar, A. Bhorkar, N. Hirose, and S. Levine, ``GNM: A general navigation model to drive any robot,'' in \emph{Proc. IEEE ICRA}, 2023.
\bibitem{vint2023} D. Shah \emph{et al.}, ``ViNT: A foundation model for visual navigation,'' in \emph{Proc. CoRL}, 2023.
\bibitem{nomad2024} A. Sridhar, D. Shah, C. Glossop, and S. Levine, ``NoMaD: Goal masked diffusion policies for navigation and exploration,'' in \emph{Proc. IEEE ICRA}, 2024.
\bibitem{exaug2023} N. Hirose, D. Shah, A. Sridhar, and S. Levine, ``ExAug: Robot-conditioned navigation policies via geometric experience augmentation,'' in \emph{Proc. IEEE ICRA}, 2023.
\bibitem{crossemb2024} J. Yang \emph{et al.}, ``Pushing the limits of cross-embodiment learning for manipulation and navigation,'' in \emph{Proc. RSS}, 2024.
\bibitem{embagn2024} N. Curtis, O. Azulay, and A. Sintov, ``Embodiment-agnostic navigation policy trained with visual demonstrations, 2024.
\bibitem{onering2024} A. Eftekhar \emph{et al.}, ``The One RING: A robotic indoor navigation generalist, 2024.
\bibitem{navdp2025} W. Cai \emph{et al.}, ``NavDP: Learning sim-to-real navigation diffusion policy with privileged information guidance,'' arXiv:2505.08712, 2025.
\bibitem{se2025} H. He, Y. Ma, B. Squicciarini, W. Wu, and B. Zhou, ``From seeing to experiencing: Scaling navigation foundation models with reinforcement learning,'' arXiv:2507.22028, 2025.
\bibitem{navila2025} A.-C. Cheng \emph{et al.}, ``NaVILA: Legged robot vision-language-action model for navigation,'' in \emph{Proc. RSS}, 2025.
\bibitem{internvlan1} M. Wei \emph{et al.}, ``Ground slow, move fast: A dual-system foundation model for generalizable vision-and-language navigation,'' arXiv:2512.08186, 2025.
\bibitem{trackvla2025} S. Wang \emph{et al.}, ``TrackVLA: Embodied visual tracking in the wild,'' in \emph{Proc. CoRL}, 2025.
\bibitem{navfom2025} J. Zhang \emph{et al.}, ``Embodied navigation foundation model,'' arXiv:2509.12129, 2025.
\bibitem{ticvla2026} Z. Huang, Y. Zhang, J. Liu, R. Song, C. Tang, and J. Ma, ``TIC-VLA: A think-in-control vision-language-action model for robot navigation in dynamic environments,'' in \emph{Proc. ICML}, 2026.
\bibitem{quarvla2024} P. Ding \emph{et al.}, ``QUAR-VLA: Vision-language-action model for quadruped robots,'' in \emph{Proc. ECCV}, 2024.
\bibitem{quartonline2024} X. Tong \emph{et al.}, ``QUART-Online: Latency-free large multimodal language model for quadruped robot learning,'' in \emph{Proc. IEEE ICRA}, 2025.
\bibitem{navid2024} J. Zhang \emph{et al.}, ``NaVid: Video-based VLM plans the next step for vision-and-language navigation,'' in \emph{Proc. RSS}, 2024.
\bibitem{uninavid2024} J. Zhang \emph{et al.}, ``Uni-NaVid: A video-based vision-language-action model for unifying embodied navigation tasks,'' in \emph{Proc. RSS}, 2025.
\bibitem{streamvln2025} M. Wei \emph{et al.}, ``StreamVLN: Streaming vision-and-language navigation via SlowFast context modeling,'' in \emph{Proc. IEEE ICRA}, 2026.
\bibitem{behav2025} K. Weerakoon \emph{et al.}, ``BehAV: Behavioral rule guided autonomy using VLMs for robot navigation in outdoor scenes,'' in \emph{Proc. IEEE ICRA}, 2025.
\bibitem{convoi2024} A. J. Sathyamoorthy \emph{et al.}, ``CoNVOI: Context-aware navigation using vision language models in outdoor and indoor environments,'' in \emph{Proc. IEEE/RSJ IROS}, 2024.
\bibitem{vltgs2025} D. Song, J. Liang, X. Xiao, and D. Manocha, ``VL-TGS: Trajectory generation and selection using vision language models in mapless outdoor environments,'' \emph{IEEE Robot. Autom. Lett.}, 2025.
\bibitem{dowhatyousay2025} Y. Wu, A. Li, T. Hermans, F. Ramos, A. Bajcsy, and C. P\'{e}rez-D'Arpino, ``Do what you say: Steering vision-language-action models via runtime reasoning-action alignment verification,'' arXiv:2510.16281, 2025.
\bibitem{codevilla2018} F. Codevilla, M. M\"{u}ller, A. L\'{o}pez, V. Koltun, and A. Dosovitskiy, ``End-to-end driving via conditional imitation learning,'' in \emph{Proc. IEEE ICRA}, 2018.
\bibitem{chaplot2018} D. S. Chaplot, K. M. Sathyendra, R. K. Pasumarthi, D. Rajagopal, and R. Salakhutdinov, ``Gated-attention architectures for task-oriented language grounding,'' in \emph{Proc. AAAI}, 2018.
\bibitem{diffdrive2025} B. Liao \emph{et al.}, ``DiffusionDrive: Truncated diffusion model for end-to-end autonomous driving,'' in \emph{Proc. IEEE/CVF CVPR}, 2025.
\bibitem{ddpg2016} T. P. Lillicrap \emph{et al.}, ``Continuous control with deep reinforcement learning,'' in \emph{Proc. ICLR}, 2016.
\bibitem{germ2024} W. Song \emph{et al.}, ``GeRM: A generalist robotic model with mixture-of-experts for quadruped robot,'' in \emph{Proc. IEEE/RSJ IROS}, 2024.
\bibitem{diffdrivev2} J. Zou \emph{et al.}, ``DiffusionDriveV2: Reinforcement learning-constrained truncated diffusion modeling in end-to-end autonomous driving,'' arXiv:2512.07745, 2025.
\bibitem{wang2020} X. E. Wang \emph{et al.}, ``Environment-agnostic multitask learning for natural language grounded navigation,'' in \emph{Proc. ECCV}, 2020.
\bibitem{gervet2023} T. Gervet, S. Chintala, D. Batra, J. Malik, and D. S. Chaplot, ``Navigating to objects in the real world,'' \emph{Science Robotics}, 2023.
\bibitem{lelan2024} N. Hirose, C. Glossop, A. Sridhar, O. Mees, and S. Levine, ``LeLaN: Learning a language-conditioned navigation policy from in-the-wild video,'' in \emph{Proc. CoRL}, 2024.
\bibitem{scand2022} H. Karnan \emph{et al.}, ``Socially compliant navigation dataset (SCAND): A large-scale dataset of demonstrations for social navigation,'' \emph{IEEE Robot. Autom. Lett.}, 2022.
\bibitem{gnd2025} J. Liang \emph{et al.}, ``GND: Global navigation dataset with multi-modal perception and multi-category traversability in outdoor campus environments,'' in \emph{Proc. IEEE ICRA}, 2025.
\bibitem{musohu2023} D. M. Nguyen, M. Nazeri, A. Payandeh, A. Datar, and X. Xiao, ``Toward human-like social robot navigation: A large-scale, multi-modal, social human navigation dataset,'' in \emph{Proc. IEEE/RSJ IROS}, 2023.
\bibitem{sacson2023} N. Hirose, D. Shah, A. Sridhar, and S. Levine, ``SACSoN: Scalable autonomous control for social navigation,'' \emph{IEEE Robot. Autom. Lett.}, 2024.
\bibitem{citywalker2025} X. Liu \emph{et al.}, ``CityWalker: Learning embodied urban navigation from web videos,'' in \emph{Proc. IEEE/CVF CVPR}, 2025.
\bibitem{navida2026} W. Zhu \emph{et al.}, ``NaVIDA: Vision-language navigation with inverse dynamics augmentation,'' arXiv:2601.18188, 2026.
\bibitem{gptdriver2023} J. Mao, Y. Qian, J. Ye, H. Zhao, and Y. Wang, ``GPT-Driver: Learning to drive with GPT,'' arXiv:2310.01415, 2023.
\bibitem{slowbrain2026} Z. Peng, H. He, Q. Li, Y. Ma, and B. Zhou, ``Slow brain, fast planner: Latency-resilient VLM-augmented urban navigation,'' arXiv:2606.20458, 2026.
\bibitem{vlmsocialnav2025} D. Song, J. Liang, A. Payandeh, A. H. Raj, X. Xiao, and D. Manocha, ``VLM-Social-Nav: Socially aware robot navigation through scoring using vision-language models,'' \emph{IEEE Robot. Autom. Lett.}, vol. 10, no. 1, pp. 508--515, 2025.
\bibitem{clipseg} T. L\"{u}ddecke and A. S. Ecker, ``Image segmentation using text and image prompts,'' in \emph{Proc. IEEE/CVF CVPR}, 2022.
\bibitem{yoloworld} T. Cheng \emph{et al.}, ``YOLO-World: Real-time open-vocabulary object detection,'' in \emph{Proc. IEEE/CVF CVPR}, 2024.
\bibitem{clip} A. Radford \emph{et al.}, ``Learning transferable visual models from natural language supervision,'' in \emph{Proc. ICML}, 2021..
\bibitem{dav2} L. Yang \emph{et al.}, ``Depth Anything V2,'' in \emph{Proc. NeurIPS}, 2024.
\end{thebibliography}
\end{document}